%% file: main.tex
\RequirePackage{fix-cm}
\documentclass{article} %
\PassOptionsToPackage{sort&compress}{natbib}
\usepackage{vista,times}

\usepackage{amsmath,amsfonts,bm}

\usepackage{listings}
\usepackage{algorithm}
\makeatletter
\newcommand\fs@vistaruled{%
  \def\@fs@cfont{\bfseries}\let\@fs@capt\floatc@ruled
  \def\@fs@pre{\hrule height.8pt depth0pt \kern5pt}%
  \def\@fs@post{\kern4pt\hrule\relax}%
  \def\@fs@mid{\kern5pt\hrule\kern5pt}%
  \let\@fs@iftopcapt\iftrue}
\makeatother
\makeatletter
\newcommand\floatc@vistanonum[2]{#2\par}
\newcommand\fs@vistanonum{%
  \def\@fs@cfont{}\let\@fs@capt\floatc@vistanonum
  \def\@fs@pre{\hrule height.8pt depth0pt \kern5pt}%
  \def\@fs@post{\kern4pt\hrule\relax}%
  \def\@fs@mid{\kern5pt\hrule\kern5pt}%
  \let\@fs@iftopcapt\iftrue}
\makeatother
\floatstyle{vistaruled}
\newfloat{promptbox}{tbp}{lop}
\floatname{promptbox}{Prompt}
\floatstyle{vistanonum}
\restylefloat{algorithm}
\usepackage{xcolor}
\usepackage{hyperref}
\hypersetup{hidelinks}
\DeclareFontShape{OT1}{cmtt}{b}{n}{<->ssub*cmtt/m/n}{}
\DeclareFontFamily{OML}{cmtt}{}
\DeclareFontShape{OML}{cmtt}{m}{n}{<->ssub*cmm/m/it}{}
\DeclareFontFamily{OMS}{cmtt}{}
\DeclareFontShape{OMS}{cmtt}{m}{n}{<->ssub*cmsy/m/n}{}

\setcitestyle{numbers,square,comma}
\usepackage{url}
\usepackage{graphicx}
\usepackage{wrapfig}
\usepackage{placeins}

\makeatletter
\newcommand{\needlines}[1]{\par\penalty-100\begingroup
  \setlength{\dimen@}{#1\baselineskip}%
  \dimen@ii\pagegoal \advance\dimen@ii-\pagetotal
  \ifdim\dimen@>\dimen@ii \ifdim\dimen@ii>\z@\vfil\fi\penalty-\@M\fi
  \endgroup}
\makeatother
\usepackage{booktabs}
\usepackage{multirow}
\title{VISTA: A Visual Harness for Reasoning\\in an Interactive World}

\definecolor{linkpink}{RGB}{210, 80, 140}
\author{%
\textbf{Qiushi Han\textsuperscript{*} \qquad Keya Hu\textsuperscript{*} \qquad Linlu Qiu\textsuperscript{*} \qquad Cathy Wu \qquad Kaiming He} \\[6pt]
Massachusetts Institute of Technology \\[6pt]
\textbf{Code:} \textcolor{linkpink}{\bfseries\url{https://github.com/joshhhhhan/VISTA}}
}

\begin{document}

\maketitle

\begingroup
\renewcommand{\thefootnote}{}
\makeatletter
\renewcommand{\@makefntext}[1]{\noindent#1}
\makeatother
\footnotetext{\textsuperscript{*}Co-first authors.
Email: \texttt{\{joshhan, keya\_hu, linluqiu, cathywu, kaiming\}@mit.edu}}
\endgroup

\input{src/00_abstract}
\input{src/01_introduction}
\input{src/02_related_work}

\input{src/03_method}

\input{src/04_arc_agi_3}

\input{src/06_analysis}

\input{src/05_beyond_arc}

\input{src/07_conclusion}

\bibliography{references}
\bibliographystyle{plainnat}

\raggedbottom
\makeatletter\setlength{\@fptop}{0pt}\makeatother
\appendix
\input{src/09_appendix}

\end{document}

%% file: src/00_abstract.tex
\begin{abstract}
We show that multimodal models possess strong reasoning abilities and that an appropriate harness can unlock their potential to solve tasks across diverse interactive environments. We introduce \textbf{VISTA}, a visual harness that gives a general-purpose multimodal model long-horizon vision. VISTA allows the model to directly perceive the environment through visual observations and maintains a lossless visual memory that preserves past observations in their original form. The model can actively retrieve these observations and reorganize its visual input as it reasons. On ARC-AGI-3, VISTA improves Claude Opus 5.0's Relative Human Action Efficiency score from 40.68 
to a perfect 100.00, with the model completing all 25 public games using 57.4\% fewer actions than first-time human participants. VISTA's simple design also allows it to extend naturally to diverse visual environments with minimal adaptation. Across three additional benchmarks covering a diverse range of visual games and puzzles, it substantially outperforms baselines using the same underlying model with minimal harnesses. Our results highlight VISTA's potential as a general-purpose visual harness for advancing multimodal agents in complex visual environments.
\end{abstract}

%% file: src/01_introduction.tex
\section{Introduction}
\label{sec:intro}

\needlines{14}
\begin{wrapfigure}{r}{0.47\linewidth}
\vspace{-1.1\baselineskip}
\centering
\includegraphics[width=\linewidth]{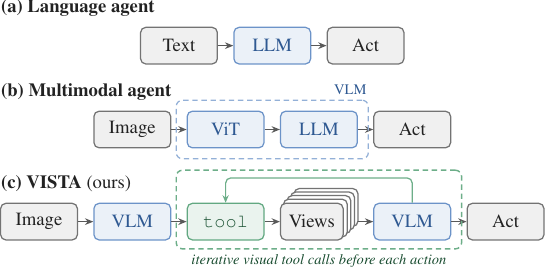}
\vspace{\dimexpr-1.1em+3.20pt\relax}
\caption{\textbf{Comparison of different agent designs.} (a) Language agents reason over textual observations. (b) Multimodal agents encode images into visual representations only once, but these representations may be compressed, lossy, and insufficient for subsequent reasoning, potentially limiting the models' reasoning capabilities. (c) VISTA allows the multimodal model to iteratively call visual tools to retrieve past visual observations and reorganize its visual input as it reasons. For simplicity, we show the VLM as a vision transformer (ViT) encoder followed by an LLM. The two VLM blocks represent the same underlying model.}
\label{fig:teaser}
\vspace{\dimexpr-2\baselineskip-4.20pt\relax}
\end{wrapfigure}

We live in a dynamic, high-dimensional world. As humans, we perceive our surroundings, reason about what we observe, and act based on our understanding and past experience~\citep{wu2024vstar,su2025thinkingwithimages}. However, it remains unclear whether existing artificial intelligence (AI) systems have a similar ability to understand and interact robustly with the world, especially in unfamiliar environments.

To study this question, we use ARC-AGI-3~\citep{arcprize2026arc3}, a suite of interactive visual games in which an agent needs to discover the rules and goals of unfamiliar worlds through interaction. This benchmark is typically approached as a language-based problem, with large language models (LLMs) receiving a textual or structured representation of the environment, as shown in Figure~\ref{fig:teaser}a. Building on this setup, a common strategy is to use code-based methods that construct a executable world model using symbolic programs~\citep{schema2026,astroseger2026ewma,brown2026retrodict,tycho2026}.

We instead reframe ARC-AGI-3 as a vision-based problem and address it using multimodal models. However, as illustrated in Figure~\ref{fig:teaser}b, existing vision-language models (VLMs) typically encode each image once into visual representations~\citep{alayrac2022flamingo,liu2023visualinstruction}. These representations may not preserve the details needed as reasoning evolves. Moreover, earlier observations may be removed from the context or replaced by textual summaries to stay within the model's context limit~\citep{liu2023lostmiddle,packer2023memgpt,liu2026atmem}. These limitations may constrain the models' reasoning capabilities during long-horizon interactions.

\begin{figure}[t]
\centering
\includegraphics[width=0.95\linewidth]{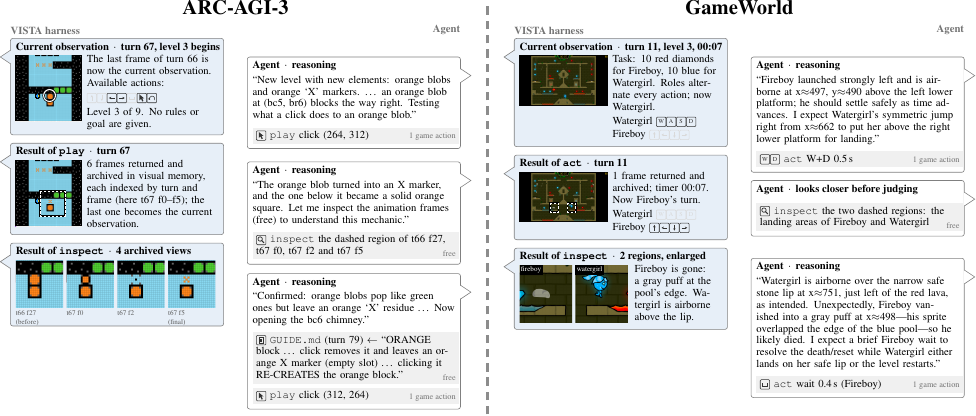}
\caption{\textbf{VISTA trajectories} on ARC-AGI-3 (left) and GameWorld
(right). The harness allows the agent to directly observe the visual states, reason in natural language, and inspect its lossless visual memory at the original level of detail when needed. }
\label{fig:trace}
\end{figure}

To this end, we introduce \textbf{VISTA}, a visual harness that allows the model to actively reorganize its visual input during long-horizon interactions (Figure~\ref{fig:teaser}c). The model receives \emph{visual observations} of the environment that preserve the appearance and spatial relations of objects. It develops its understanding of the environment through free-form language reasoning~\citep{wei2022chain,shinn2023reflexion} and uses this understanding to predict action outcomes and plan subsequent actions~\citep{yao2023react}.
Instead of relying solely on visual representations retained in the model's context, which may be compressed, lossy, and insufficient for subsequent reasoning, VISTA maintains a \emph{lossless visual memory}. This memory stores every frame returned by the environment, including intermediate animation frames, in its original form. The model can retrieve observations from this memory through \emph{visual inspection}, selecting frames to compare and spatial regions to examine at different levels of detail based on its current reasoning needs~\citep{su2025thinkingwithimages,wang2025pixelreasoner}.
Figure~\ref{fig:trace} shows two example agent trajectories with VISTA, illustrating how observation, reasoning, inspection, and action are interleaved. 

With VISTA, Claude Opus 5.0~\citep{anthropic2026opus5} completes all 25 public ARC-AGI-3 games with a Relative Human Action Efficiency (RHAE) score of 100.00 with xhigh reasoning effort, using 57.4\% fewer actions than the first-time human baseline. VISTA with GPT-5.6 Sol~\citep{openai2026gpt56} likewise completes all 25 games, achieving an RHAE score of 99.00 with max reasoning effort. To our knowledge, VISTA is the first vision-based system to achieve perfect or near-perfect performance on ARC-AGI-3 without program synthesis. For comparison, the organizers' official baselines, which also do not use programs, achieve RHAE scores of 40.68 
with Opus 5.0 high and 13.33 with GPT-5.6 Sol max~\citep{arcprize2026gpt56results,openai2026twosettings}. 

The simplicity of VISTA also allows it to extend naturally to other visual environments with minimal adaptation. We apply the same visual harness to 34 browser games from GameWorld~\citep{ouyang2026gameworld}, 10 browser games from AI GameStore~\citep{ying2026aigamestore}, and 39 visual tracking problems from BabyVision~\citep{chen2026babyvision}. Together, these benchmarks cover a wide range of games and visual reasoning tasks that require skills such as complex reasoning, spatial navigation, and strategic planning. GPT-5.6 Sol with VISTA substantially outperforms baselines that use the same model with minimal harnesses. It also outperforms the novice human baseline on GameWorld and the human median baseline on AI GameStore. Overall, these results demonstrate VISTA's effectiveness across diverse visual environments and highlight visual harness design as an important direction for improving multimodal agents.

%% file: src/02_related_work.tex
\section{Related Work}
\label{sec:related}

\paragraph{ARC-AGI benchmarks.}

ARC-AGI-1~\citep{chollet2019measure} and ARC-AGI-2~\citep{chollet2025arc2} are benchmarks designed to evaluate abstract reasoning over static input--output grids~\citep{arcprize2026techreport}. Approaches to these benchmarks often use pretrained large language models~\citep{berman2024record536,akyurek2024surprising,franzen2025productexpertsllmsboosting,puget2024arc,li2024combining,berman2025highestscore}, recurrent models~\citep{wang2025hierarchicalreasoningmodel,jolicoeurmartineau2025morerecursivereasoningtiny}, or vision-based models~\citep{hu2025arcvision}.
ARC-AGI-3~\citep{arcprize2026arc3} makes this setting interactive: an agent must explore novel environments, adapt continuously from its interactions, and take actions to achieve its goals~\citep{legris2024harc}. Existing approaches to ARC-AGI-3 largely rely on program synthesis~\citep{schema2026,tycho2026,brown2026retrodict,astroseger2026ewma,rodionov2026execworldmodels}, in which agents perceive the environment through textual or symbolic representations and abstract it into programs that can be executed, verified, and searched over to guide actions. While effective on ARC-AGI-3, these approaches rely on constructing explicit programmatic abstractions of the environment. As environments become more visually complex, representing their states, transition dynamics, and goal conditions with explicit programs may become increasingly difficult. In contrast, VISTA operates directly on visual observations and reasons in natural language.

\vspace{-0.5em}
\paragraph{Computer-use agents.}
Recent work has explored agents that interact with computers and graphical user interfaces to complete complex tasks. Existing approaches broadly differ in their observation space. \emph{Text-based} approaches operate over structured representations of the interface, such as HTML, DOM trees, accessibility trees, or element identifiers, and use language models to reason over these representations and take symbolic actions~\citep{nakano2021webgpt,deng2023mind2web,zhou2023webarena}. \emph{Vision-based} computer-use agents interact directly with screenshots and take grounded actions such as clicks, mouse movements, and keyboard inputs~\citep{cheng2024seeclick,xu2025aguvis,qin2025ui,wang2025opencua,lu2026qwencua}. VISTA follows the vision-based paradigm: the agent reasons directly from visual observations rather than textual or symbolic representations of the environment. It focuses on a visual-centric harness for reasoning in interactive worlds and provides a simple interface for agents to observe and interact with the environment.

\vspace{-0.5em}
\paragraph{Game-playing agents.}
Games have long served as testbeds for developing agents that learn to perceive, reason, and act in interactive environments. Classical approaches learn policies through reinforcement learning~\citep{mnih2015human,silver2016mastering,silver2017mastering} or learn environment dynamics to support prediction and planning~\citep{schrittwieser2020mastering,hafner2023dreamerv3}. While these methods provide general learning algorithms, they typically require training a separate agent for each game. More recently, foundation models have enabled more general-purpose agents that operate in richer game environments. Voyager uses a large language model together with structured Minecraft observations and an executable skill library~\citep{wang2023voyager}, while JARVIS-1 incorporates visual observations and retrieves similar past experiences for planning in Minecraft~\citep{wang2023jarvis1}. Generalist agents such as SIMA and SIMA~2~\citep{raad2024sima,bolton2025sima2}, Cradle~\citep{tan2025cradle}, and Game-TARS~\citep{wang2025gametars} further operate directly from visual observations across multiple environments. A central challenge in these systems is retaining information over long interaction horizons: prior agents typically maintain only a bounded visual context or selectively store and compress past experiences. VISTA instead preserves the complete visual interaction history losslessly and allows the agent to actively revisit arbitrary previous frames and regions during reasoning.

%% file: src/03_method.tex
\section{VISTA: A Visual Harness}
\label{sec:method}

\begin{figure}[t]
\centering
\includegraphics[width=\linewidth]{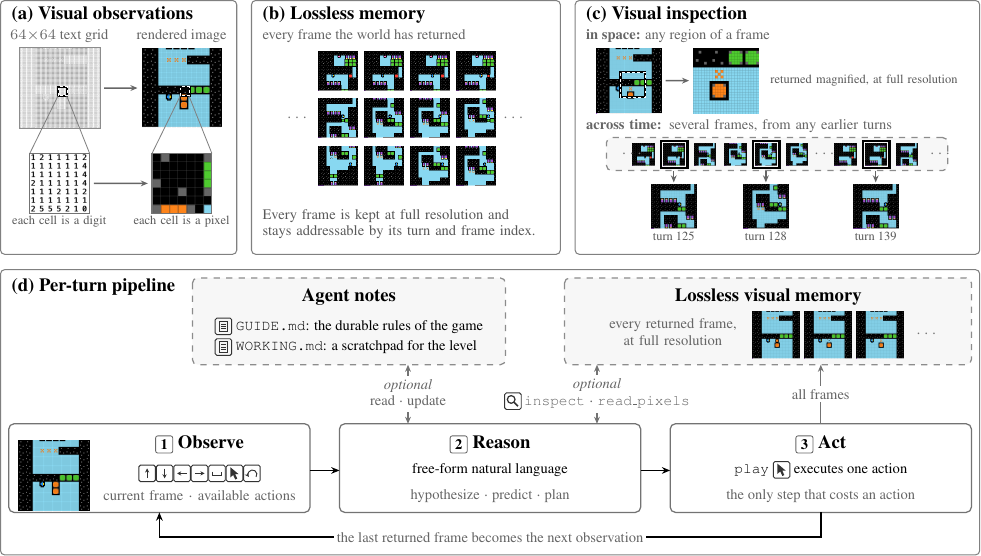}
\caption{\textbf{VISTA's key design components and per-turn pipeline.}
We use ARC-AGI-3 as an example.
(a) \textit{Visual observations}: The model receives rendered visual states.
(b) \textit{Lossless visual memory}: Every returned frame is stored in its original form.
(c) \textit{Model-directed visual inspection}: The model selects frames and regions to examine during reasoning.
(d) \textit{Per-turn pipeline}: The agent observes, reasons, and acts, while each returned frame is preserved in visual memory. Revisiting past observations and updating notes are optional steps directed by the model.}
\label{fig:designs}
\end{figure}

VISTA lets the model directly perceive the environment through visual inputs. However, existing multimodal systems typically encode visual observations once and discard the raw inputs. As the context fills, these representations may be removed or replaced by textual summaries, and the model may not control what is retained. More importantly, the model may not know at encoding time which visual information will matter later: a representation sufficient at one stage may omit details required in subsequent reasoning steps.

These limitations directly motivate the design of VISTA, which preserves original observations in a lossless visual memory and provides tools for the model to retrieve and inspect them during interaction. This allows the model to revisit visual evidence and reorganize its visual input as it reasons, without requiring all relevant information to be captured in the initial visual encoding and retained in the model's context.

\subsection{Visual Harness Components}
\label{sec:components}

We introduce three key visual harness components: \emph{visual observations}, which allows the agent to perceive the environment through rendered images; \emph{lossless visual memory}, which preserves visual observations in their original form; and \emph{visual inspection}, which allows the model to revisit visual evidence and examine details as needed. Together, these components provide the model with an \emph{explicit attention mechanism} over the interaction history, allowing it to select which frames and regions to examine during reasoning.

\vspace{-0.5em}
\paragraph{Visual observation.}
\label{sec}

The model receives visual observations from the environment (Figure.~\ref{fig:designs}a). These observations preserve the appearance and spatial relation of objects, allowing the model to use its learned visual priors to interpret and reason about the environment.

\vspace{-0.5em}
\paragraph{Lossless visual memory.}
\label{sec:memory}
The harness stores every frame returned by the environment, including intermediate animation frames, in its original form (Figure.~\ref{fig:designs}b). Each frame is indexed by turn and frame number and remains available throughout the interaction, independently of what is retained in the model's current context.

\vspace{-0.5em}
\paragraph{Visual inspection.}

Building on the visual memory, we provide the agent with tools to actively inspect visual information in space and across time (Figure~\ref{fig:designs}c). \emph{In space}, it can request an enlarged view of any rectangular region from the original frame, allowing it to examine fine-grained details that may be difficult to perceive in the full image. \emph{Across time}, it can retrieve and inspect multiple frames from arbitrary earlier turns to revisit past visual evidence, compare states, and replay how an action unfolded through its intermediate animation frames. At each turn, the model is free to decide whether to inspect its visual memory and which frames or regions to examine.

\subsection{Pipeline}
\label{sec:pipeline}

In our pipeline, the model and the harness interact through a multi-turn conversation (Figure~\ref{fig:trace}). The harness is purely mechanical and does not involve any trained model: it executes tool calls requested by the model, returns outputs such as images or pixel values, and reports errors for invalid requests.
The model reasons over these outputs and issues its next request to the harness if needed.

A single agent interacts with each game throughout the run. Each game turn may include multiple exchanges with the harness before the agent executes one game action (Figure~\ref{fig:designs}d). At the start of a turn, the agent receives the current visual state and available actions. It can also consult two notes that it maintains: \texttt{GUIDE.md}, which records its understanding of the game and information that may remain useful across levels, and \texttt{WORKING.md}, which serves as a scratchpad for the current state. Through \texttt{inspect}, the agent can revisit earlier observations, including intermediate animation frames produced by its previous actions. It can also use \texttt{read\_pixels} to examine pixel values within a selected region. Consulting the notes and using the inspection tools are optional and left to the agent's decision.

The agent uses free-form language to reason and plan. It predicts the outcome of its planned action, revises its notes to reflect its updated understanding of the environment, and calls \texttt{play} to execute the chosen action and advance to the next game turn. When its context approaches the limit, the agent writes a concise handoff summary and resumes from the current state in a fresh context. Its notes (\texttt{GUIDE.md} and \texttt{WORKING.md}), visual memory, and action history remain available.

Every game uses the same interface and short prompt as shown in Appendix~\ref{app:prompt}. The available tools (e.g., \texttt{inspect}, \texttt{read\_pixels}, and \texttt{play}), together with their pseudocode and corresponding descriptions are provided to the agent, as shown in Appendix~\ref{app:tool_prompt}).

%% file: src/04_arc_agi_3.tex
\section{VISTA on ARC-AGI-3}
\label{sec:arc}

\subsection{Setup}

\begin{figure}[t]
\centering
\includegraphics[width=\linewidth]{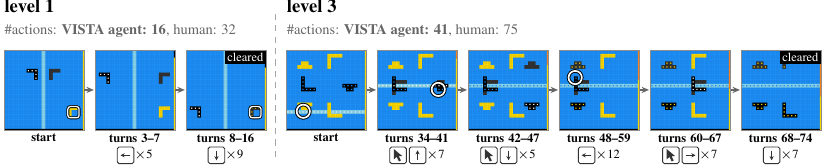}
\caption{\textbf{Example trajectory in ARC-AGI-3.} Each frame is labeled with its turn and action; $\times N$ denotes repeated actions. Boxes mark targets, and rings mark the agent's clicks. The agent must infer the rules: black pieces move together with their reflections across the mirror line, and all yellow targets must be covered. Level~3 introduces a movable horizontal mirror line and selectable pieces. 
}
\label{fig:arc3-traj}
\end{figure}

\paragraph{Dataset.}
ARC-AGI-3~\citep{arcprize2026arc3} consists of interactive visual games whose rules and objectives are not provided explicitly. We evaluate on the 25 public games. At each step, the agent observes the current visual state, selects an available action, and uses the resulting observation to infer object roles, transition rules, hidden states, and possible goals. Each game contains multiple levels that share underlying mechanics while introducing new concepts, objects, layouts, or constraints. The agent therefore needs to retain and apply previously inferred knowledge across levels while continually adapting its understanding through interaction. 
Figure~\ref{fig:arc3-traj} shows an example agent trajectory.

\vspace{-0.5em}
\paragraph{Evaluation metrics.}
Performance is measured by Relative Human Action Efficiency (RHAE)~\citep{arcprize2026methodology}, which compares an agent's action efficiency with that of first-time human players. A completed level receives 100\% credit when its action count matches the human baseline, up to 115\% when it uses fewer actions, and partial credit when it uses more actions. Unfinished levels receive zero credit. Later levels receive greater weight, and each game score is capped by the weighted fraction of levels completed. See Appendix~\ref{app:rhae} for details.

\vspace{-0.5em}
\paragraph{Models.}

We instantiate VISTA with two general-purpose, closed-weight multimodal model backends commonly used in prior work. Each backend is evaluated independently under the same interaction protocol.
We use Claude Opus 5.0~\citep{anthropic2026opus5} run with extra-high reasoning effort through the Claude Code CLI and GPT-5.6 Sol~\citep{openai2026gpt56} run with max reasoning effort through the Codex CLI.

\subsection{Results}
\label{sec:arc-results}

\input{tables/system_level_comparison}
Table~\ref{tab:arc} shows a system-level comparison on the 25 public ARC-AGI-3 games, unless otherwise noted. With Claude Opus 5.0, VISTA achieves a perfect RHAE score of 100.00, matching other state-of-the-art systems. It uses 7,302 actions in total, 57.4\% fewer than the human reference of 17,135 actions.

Most of the compared systems come from concurrent work. These systems typically represent observations as text-based numerical grids and often reason using a code-based world model: the agent abstracts the environment into symbolic programs that can be executed and checked during exploration. 
To our knowledge, VISTA is the first system to achieve perfect or near-perfect performance on ARC-AGI-3 \emph{without program synthesis}. For comparison, the same model scores only 40.68  under the official implementation.
In Figure~\ref{fig:repr}, we also compare textual grids with our default image observations. We find that visual observations achieve comparable performance with fewer actions while using significantly fewer tokens.

%% file: tables/system_level_comparison.tex
\begin{wraptable}{r}{0.5\linewidth}
\vspace{-1.2\baselineskip}
\caption{\textbf{System-level comparison} on 25 public games of ARC-AGI-3. VISTA is, to our knowledge, the first system to achieve perfect or near-perfect performance without program synthesis.}
\vspace{0.5em}
\label{tab:arc}
\centering
\fontsize{7}{8}\selectfont
\setlength{\tabcolsep}{3pt}
\renewcommand{\arraystretch}{1.15}
\begin{tabular}{llllr}
\toprule
\textbf{System} & \textbf{Program} & \textbf{Model} & \textbf{Effort} & \textbf{RHAE} \\
\midrule
\midrule
\multirow{2}{*}{Official impl.~\citep{arcprize2026arc3}}
 & \multirow{2}{*}{No} & GPT-5.6 Sol~\citep{openai2026twosettings} & max & 13.33 \\
 & & Opus 5.0$^{\dagger}$ & high & 40.68 \\
\midrule
\multirow{2}{*}{Schema~\citep{schema2026}}
 & \multirow{2}{*}{Yes} & GPT-5.6 Sol & xhigh $\rightarrow$ max & 95.35 \\
 & & Opus 4.8 $\rightarrow$ Fable 5 & max & 98.98 \\
\midrule
ewma\_sv\_v1.6~\citep{astroseger2026ewma}
 & Yes & GPT-5.6 Sol & xhigh & 98.97 \\
\midrule
Retrodict~\citep{brown2026retrodict}
 & Yes & GPT-5.6 Sol & max & 99.86 \\
\midrule
\multirow{2}{*}{Tycho~\citep{tycho2026}}
 & \multirow{2}{*}{Yes} & GPT-5.6 Sol & max & \textbf{100.00} \\
 & & Opus 5.0 & xhigh & \textbf{100.00} \\
\midrule
\midrule
\multirow{2}{*}{\textbf{VISTA}}
 & \multirow{2}{*}{\textbf{No}} & GPT-5.6 Sol & max & 99.00 \\
 & & Opus 5.0 & xhigh & \textbf{100.00} \\
\bottomrule
\end{tabular}
\par\vspace{3pt}\raggedright{\fontsize{6.5pt}{7.5pt}\selectfont $^{\dagger}$Calculated from the officially reported per-game scores for the 25 public games~\citep{arcprize2026opus5results}.\par}
\vspace{\baselineskip}
\end{wraptable}

%% file: src/06_analysis.tex
\subsection{Ablations}
\label{sec:ablations}

For all ablations, we use GPT-5.6 Sol with max reasoning effort and a 200K-token context limit as the default setting. The agent receives a $512\times512$ PNG image of the current rendered state, obtained by $8\times$ nearest-neighbor upscaling of the official $64\times64$ frame.

\begin{figure}[t]
\centering
\includegraphics[width=\linewidth]{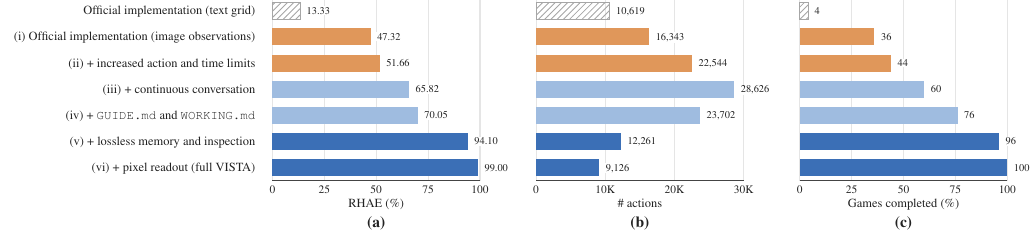}%
\caption{\textbf{Harness configurations on ARC-AGI-3.} GPT-5.6 Sol under progressively richer harness configurations, evaluated using (a) RHAE, (b) the total number of actions taken across the 25 games, and (c) the percentage of the 25 games completed. For reference, the official implementation using textual grids (hatched gray bars) achieves a reported RHAE score of 13.33 with 10,619 actions~\citep{arcprize2026gpt56results,openai2026twosettings}. (i) The official implementation with PNG images replacing its textual grid observations. (ii) Increases the action and time limits to match our evaluation protocol (up to 2,000 actions per game) instead of the official stopping criteria. (iii) Adds a continuous conversation, using the model's native compaction to continue past its context limit. (iv) Adds the two notes, \texttt{GUIDE.md} and \texttt{WORKING.md}, and encourages the agent to build a compact and revisable model of the game. (v) Adds lossless visual memory and inspection. (vi) Adds exact pixel readout, completing VISTA. Configurations (i)--(iv) automatically receive up to seven frames per action, following the official protocol; configurations (v)--(vi) introduce VISTA's key visual components: the agent automatically receives only the final frame and can actively inspect its visual memory as it reasons. See Appendix~\ref{app:text-vs-image} for details.}
\label{fig:harness-steps}
\end{figure}

\vspace{-0.5em}
\paragraph{Effects of VISTA components.}
We compare GPT-5.6 Sol under progressively richer harness configurations in Figure~\ref{fig:harness-steps}. The official implementation uses textual grids and has a reported RHAE score of 13.33 with 10,619 actions~\citep{arcprize2026gpt56results,openai2026twosettings}. Replacing these textual grids with rendered PNG images (i) increases the score to 47.32, demonstrating the importance of visual observations. The low action count is partly due to runs reaching the action or time limit before completing the games, as suggested by the low completion rate in this configuration. Increasing the action and time limits improves the score to 51.66 (ii). 
In (iii), we introduce a minimal agent harness that maintains a continuous conversation and uses native compaction to continue beyond its context limit, improving the score to 65.82. Configuration (iv) adds two note files, \texttt{GUIDE.md} and \texttt{WORKING.md}, along with instructions encouraging the agent to build a compact and revisable model of the game, further improving the score to 70.05 (see the prompt in Appendix~\ref{app:prompt}). Finally, configurations (v)--(vi) introduce VISTA's key visual harness components: adding lossless visual memory and model-directed inspection (v) improves the score to 94.10, while exact pixel readout (vi), which provides numerical RGB values for selected pixels, completes the VISTA harness and increases the score to 99.00.

\vspace{-0.5em}
\begin{figure}[t]
\centering
\includegraphics[width=\linewidth]{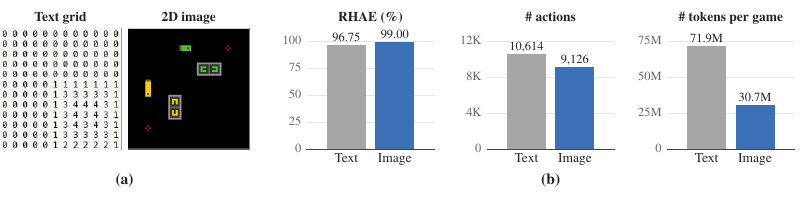}
\vspace{-1.5em}
\caption{\textbf{Comparison of observation representations on ARC-AGI-3.} (a)~An example game state shown as a flattened text grid and a 2D image; the displayed text grid is cropped from the full $64\times64$ grid. (b)~RHAE, total actions across the 25 games, and average token usage per game for the two observation representations.}
\label{fig:repr}
\end{figure}
\paragraph{Reasoning directly from images.}\label{sec:repr}

Prior approaches to ARC-AGI-3 commonly represent each frame as a textual grid, encoding observations as rows of integer color indices~\citep{schema2026,astroseger2026ewma,brown2026retrodict,tycho2026}. In contrast, VISTA reasons directly from rendered images. We ablate this design choice by replacing VISTA's image observations with textual grids while keeping the rest of the harness unchanged. As shown in Figure~\ref{fig:repr}, using textual grids maintains comparable performance but slightly increases the total number of actions. Textual grids are also substantially more token-intensive: a $64\times64$ grid requires roughly 4,000 tokens, compared with approximately 308 image tokens for our default $512\times512$ observation. Token usage per game is 71.9M with textual grids and 30.7M with image observations.

\begin{figure}[t]
\centering
\includegraphics[width=\linewidth]{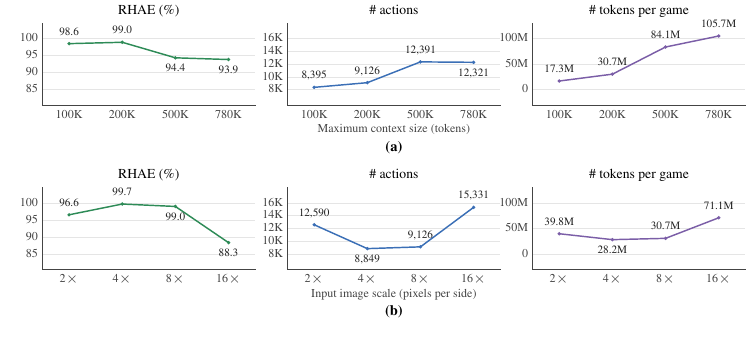}
\vspace{-2.7em}
\caption{\textbf{Ablations on ARC-AGI-3.} \textit{Left column}: RHAE. \textit{Middle column}: the total number of actions taken across the 25 games. \textit{Right column}: the average number of tokens per game.
(a): maximum context size, which sets when context compaction occurs. (b): input image scale relative to the official $64\times64$ frame; scales of $2\times$, $4\times$, $8\times$, and $16\times$ correspond to 128, 256, 512, and 1{,}024 pixels per side and roughly 20, 77, 308, and 1{,}229 image tokens per frame, respectively.}
\label{fig:ablations}
\end{figure}

\paragraph{Effect of maximum context size.}
We vary VISTA’s maximum context size, which determines how much interaction history is retained before the agent compacts its history and resumes in a fresh context. As shown in Figure~\ref{fig:ablations}a, the default 200K setting achieves the highest RHAE of 99.00. Larger contexts are substantially more expensive: 780K consumes 105.7M tokens per game, compared with 30.7M at 200K. Reducing the context to 100K lowers RHAE by only 0.4 while roughly halving token usage, showing that the harness remains effective even with shorter contexts.

\paragraph{Effect of input image scale.}
We vary the input image scale to study the trade-off between visual resolution and image token usage. The $2\times$, $4\times$, $8\times$, and $16\times$ scales correspond to 128, 256, 512, and 1,024 pixels per side, requiring approximately 20, 77, 308, and 1,229 image tokens per frame, respectively. As shown in Figure~\ref{fig:ablations}b, $4\times$ uses $75$\% fewer image tokens than the default $8\times$ while achieving comparable performance. Reducing the scale to $2\times$ degrades performance, while increasing it to $16\times$ provides no additional benefit and costs more tokens per game, likely because the extra image tokens consume context without improving visual reasoning.

\needlines{12}
\begin{wrapfigure}{r}{0.5\linewidth}
\vspace{-0.1\baselineskip}
\centering
\includegraphics[width=\linewidth]{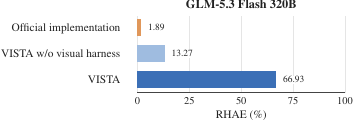}
\vspace{-1.2em}
\caption{\textbf{Results on ARC-AGI-3 using the open-weight GLM-5.3 Flash 320B model.} We compare VISTA against the official implementation using a minimal harness, as well as a VISTA variant with \texttt{inspect} and \texttt{read\_pixels} removed. VISTA's visual harness leads to a significant performance improvement.}
\label{fig:arc-glm}
\vspace{-3\baselineskip}
\end{wrapfigure}

\vspace{-0.5em}
\paragraph{Results with an open-weight model.}
\label{app:glm}
We additionally evaluate VISTA with the open-weight backend GLM-5.3 Flash 320B~\citep{glm5team2026glm5,zai2026glm53flash}. This gives a fully open-source and reproducible system. Figure~\ref{fig:arc-glm} compares the RHAE scores of three configurations: the organizers' official implementation, which we ran with its minimal harness; VISTA without the visual harness, which removes only \texttt{inspect} and \texttt{read\_pixels} tools; and full VISTA.
VISTA achieves an RHAE score of 66.93, substantially outperforming the official baseline. These results further support VISTA's effectiveness across model backends.

\begin{figure}[t]
\centering
\includegraphics[width=\linewidth]{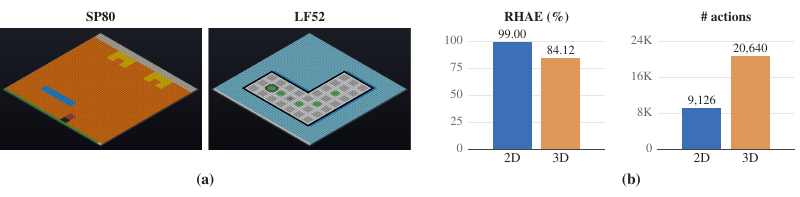}
\vspace{-1.5em}
\caption{\textbf{3D renderings of ARC-AGI-3.} (a)~Two games rendered in 3D. (b)~RHAE scores and total action counts for VISTA using 2D images and 3D renderings.}
\label{fig:repr3d}
\end{figure}

\paragraph{Effects of 3D rendering.}
We evaluate VISTA under more challenging visual conditions by rendering the same games in 3D. As shown in Figure~\ref{fig:repr3d}, the model achieves an RHAE score of 84.12 with 20,640 actions. Although VISTA requires substantially more interaction with these renderings than with the default 2D images, it still achieves strong performance and outperforms official textual baselines by a large margin. These results demonstrate its robustness to more complex visual representations and its ability to generalize beyond 2D representations. As we further show in Section~\ref{sec:gameworld}, VISTA remains effective across a broader range of visual environments, including 3D games.

\subsection{Qualitative Analysis}
\label{sec:reattention}

\begin{figure}[t]
\includegraphics[width=\linewidth]{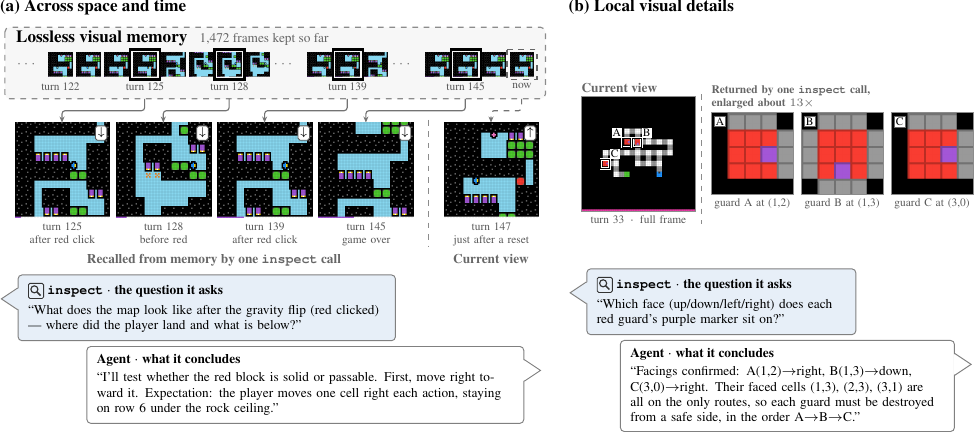}%
\caption{\textbf{Model-directed use of visual memory.}
(a) Inspect across space and time: after the board resets, the model retrieves four frames from the preceding 22 turns in a single \texttt{inspect} call to reconstruct the map.
(b) Inspect local visual details: the model uses \texttt{inspect} to zoom in on three small blocks by about $13\times$. The enlarged views reveal which side of each block carries a purple marker, allowing the model to identify each block's orientation.}
\label{fig:mem-time}
\end{figure}

The key components of the VISTA harness are lossless visual memory and visual inspection, which allow the agent to revisit stored visual information at any point during interaction. These components provide a form of explicit attention over interaction history. Here, we show qualitative examples of how agents benefit from this re-attention mechanism.
Figure~\ref{fig:mem-time}a illustrates re-attention across space and time. The agent revisits earlier frames to recover information that is no longer visible in its current observation. In this example, it inspects frames from different points in the trajectory and integrates their information to reconstruct the overall map.
Figure~\ref{fig:mem-time}b illustrates re-attention to local visual details. The agent selectively zooms into a small region of a stored observation to determine a block's orientation, which is difficult to see clearly in the full $512\times512$ frame.
Together, these examples illustrate two complementary forms of re-attention enabled by lossless visual memory: integrating visual evidence across the interaction history and inspecting local details at the granularity required by the current task.

%% file: src/05_beyond_arc.tex
\section{General Visual Environments}
\label{sec:gameworld}

\begin{figure}[t]
\centering
\setlength{\parskip}{0pt}%
\includegraphics[width=\linewidth]{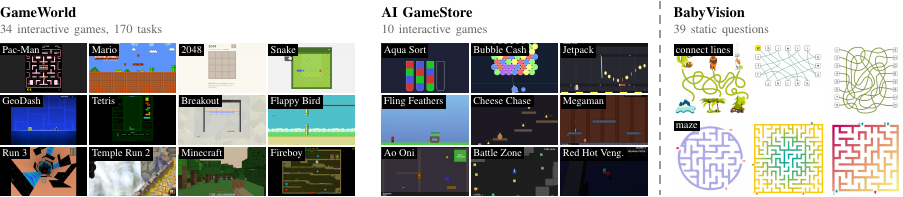}
\par\nointerlineskip\vskip 2.2mm
\includegraphics{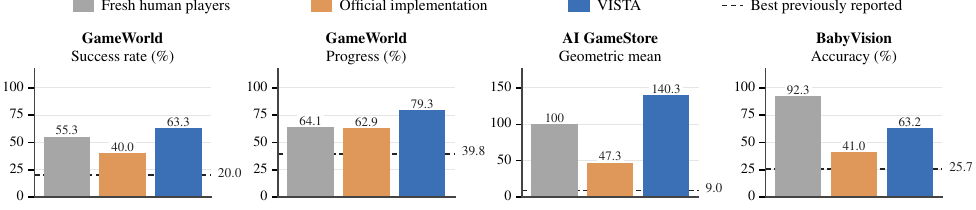}
\par
\caption{\textbf{Evaluation environments and results.} \textbf{Top}: examples from the three benchmarks. GameWorld and AI GameStore are interactive, while BabyVision is static.
\textbf{Bottom}: VISTA with GPT-5.6 Sol compared with human players and the official implementation using a minimal harness. We report success rate and progress for GameWorld, the geometric mean of human-normalized scores (human median $=100$) for AI GameStore, and accuracy for BabyVision. The dashed line marks the best previously reported result.}
\label{fig:generalization}
\end{figure}

VISTA's general-purpose design allows us to naturally apply the same visual harness to other benchmarks with minimal adaptation. We evaluate VISTA on two interactive game benchmarks: GameWorld~\citep{ouyang2026gameworld} and AI GameStore~\citep{ying2026aigamestore}. We also evaluate on a static visual reasoning benchmark, BabyVision~\citep{chen2026babyvision}, to examine the role of visual inspection without interaction with a dynamic environment.
For each benchmark, we compare VISTA with the official implementation using the same model. We also report human performance and the best previously reported result as references. See Appendices~\ref{app:rhae} and~\ref{app:human} for details. Figure~\ref{fig:generalization} shows an overview of these environments. Appendix Figure~\ref{fig:beyond} shows example VISTA trajectories on both interactive and static tasks.

\subsection{Setup}

\paragraph{GameWorld.}
GameWorld~\citep{ouyang2026gameworld} contains 34 browser games with 170 tasks across five genres: arcade, platformer, puzzle, runner, and simulation. The agent receives only the official task prompt, available controls, and screenshots. We evaluate VISTA using the computer-use interface, through which the agent interacts with the games directly via keyboard and mouse actions. 
We report success rate (SR), defined as the fraction of the 170 tasks completed, and progress (PG), which also credits partial completion. Both metrics are computed by the benchmark's official evaluator.

\vspace{-0.5em}
\paragraph{AI GameStore.}
AI GameStore~\citep{ying2026aigamestore} contains 10 public browser games. The agent receives only the public game description, controls, and visual observations. Through interaction, it must infer the game dynamics and determine how to achieve the objectives.
Each game uses its own scoring system. Following the benchmark's official protocol, we normalize each raw score so that the corresponding published human median equals 100, then compute the geometric mean of the normalized scores across the ten games.

\vspace{-0.5em}
\paragraph{BabyVision.}
BabyVision~\citep{chen2026babyvision} contains 388 static visual reasoning tasks across 22 subclasses. Each task consists of a single static image and a question. The agent perceives the image and can decide which visual regions to inspect and which individual pixels to read. We evaluate on the 39 questions from two Visual Tracking subtypes: 20 maze tasks and 19 connecting-lines tasks. We report accuracy across these tasks.

\subsection{Results}

Figure~\ref{fig:generalization} shows our results. Across all three benchmarks, VISTA consistently outperforms the corresponding baselines with minimal harness. VISTA also exceeds the reported performance of fresh human players on GameWorld and AI GameStore. On BabyVision, where observations are static and no interaction history is involved, VISTA improves accuracy from 41.0\% with the official implementation to 63.2\%, highlighting the benefit of visual inspection, which allows the agent to selectively inspect regions within a single image.

%% file: src/07_conclusion.tex
\section{Limitations}

We acknowledge that the models' existing capabilities remain a key factor in this success, and VISTA provides a simple but effective way to elicit them~\citep{yang2024sweagent,fan2026screenshots,yu2026openforgerl}. Its substantial gains over the baselines highlight visual harness design as a direction that complements model improvements. Additionally, because the models we use were released after the public benchmarks, we cannot rule out the possibility that these games were included in their training data. Evaluation on novel games, such as those in the private set of ARC-AGI-3~\citep{arcprize2026arc3}, would provide a stronger test of our system's generalization.

\section{Conclusion}
\label{sec:conclusion}

VISTA shows that a simple visual harness can unlock the potential of existing multimodal models to complete complex tasks in interactive visual environments. It provides the model with raw visual observations, preserves them in a lossless visual memory, and allows the model to revisit past visual experience at the original level of detail during interaction. We show that this simple design is sufficient for the agent to complete all 25 public ARC-AGI-3 games with a perfect score. The same harness extends naturally to other environments covering a diverse set of visual games and puzzles, with minimal adaptation.
Overall, VISTA's simple design and strong performance across diverse benchmarks suggest its potential as a general-purpose harness for tasks beyond games. Future work could evaluate VISTA on more complex and realistic tasks, such as embodied tasks in environments that more closely resemble the physical world~\citep{brohan2023rt2}.

\section*{Acknowledgments}
We thank Itamar Pres and Alan Sun for their helpful feedback and insightful discussions, and Yung-Sung Chuang and Zhaofeng Wu for their generous support.

%% file: src/09_appendix.tex
\newpage
\section{Experimental Details}
\label{app:method}

\subsection{Baselines: Official Implementations}
\label{app:baselines}

For each benchmark, we use the authors' reported results when available. Otherwise, we rerun the official implementation following the reported agent configuration. We use the models through their subscription-based CLI interfaces (Claude Opus 5.0 through the Claude Code CLI and GPT-5.6 Sol through the Codex CLI) rather than through direct API calls. For the open-weight GLM-5.3 Flash 320B , we also use the Codex CLI, with GLM as the model backend.

\paragraph{ARC-AGI-3.} On ARC-AGI-3, the organizers' minimal agent interface represents each observation as a textual grid and provides a list of available actions. When an action produces multiple frames, it retains up to seven intermediate states. It maintains a 175K-token context window; once the window is exceeded, the earliest turns are discarded, so the agent has no long-term memory. We limit the number of actions per game to 2,000 for Claude Opus 5.0 and GPT-5.6 Sol, and to 3,000 for GLM-5.3 Flash 320B.
Table~\ref{tab:arc} reports the organizers' published results for Claude Opus 5.0 and GPT-5.6 Sol~\citep{arcprize2026opus5results,arcprize2026gpt56results,openai2026twosettings}. Figure~\ref{fig:arc-glm} shows our results for GLM-5.3 Flash 320B, obtained using the original codebase.

\paragraph{GameWorld.} On GameWorld~\citep{ouyang2026gameworld}, we reproduce the official computer-use agent with GPT-5.6 Sol, following the official interaction and evaluation protocols. The agent operates directly from screenshots using keyboard and mouse actions, with each model request including the screenshots, reasoning, and actions from the two preceding steps. Each task is limited to 100 actions.

\paragraph{AI GameStore.} On AI GameStore~\citep{ying2026aigamestore}, we reproduce the official evaluation loop. The agent makes one decision per game second, receiving five frames sampled at 0.2\,s intervals. Each model request includes only these current frames as visual input, together with the scratchpad, action, and result from the preceding decision. A run has 120 decisions, with the game paused during model inference.

\paragraph{BabyVision.} On BabyVision~\citep{chen2026babyvision}, the baseline is the official one-shot setting: the model receives the image and question in a single request and answers without additional visual inspection or interaction. We extract the final answer using the benchmark's reference code.

\subsection{Ablation Details}
\label{app:text-vs-image}

We incrementally modify the official ARC-AGI-3 implementation to examine the contributions of VISTA's components (Figure~\ref{fig:harness-steps}). The official implementation accesses the model through the API, whereas our experiments use the Codex CLI through the subscription interface. Therefore, the officially reported result should be treated as an external reference. However, we use the same underlying model and reasoning effort to match the official configuration as closely as possible. All configurations (i)--(vi) use the same model access interface and are directly comparable with one another.

\textbf{(i): Image observations.}
We first replace the textual grid observations with rendered images. This leads to an RHAE score of 47.32, demonstrating the importance of visual observations. We hypothesize that improved token efficiency contributes to this gain: each image observation requires approximately 308 image tokens, compared with roughly 4K text tokens for an equivalent $64\times64$ textual grid. A request containing seven frames therefore uses roughly 28K tokens for textual observations, compared with about 2.2K for images. Image observations leave more of the model's limited context window available for reasoning within each request.

\textbf{(i)$\rightarrow$(ii): Increased action and time limits.}
The official implementation uses a per-section action limit of $5\times$ the human baseline and a 12-hour runtime limit. We find that many runs terminate upon reaching these limits, leaving the games unfinished. We therefore increase the limits to 2,000 total actions and 48 hours per game, improving the RHAE score to 51.66. We do not observe substantial additional benefits from further increasing these limits.

\textbf{(ii)$\rightarrow$(iii): Continuous conversation.}
We next allow the agent to interact with the environment through a continuous conversation. In the official implementation, each model request produces at most one environment action, and the interaction history is provided again with the next request. When this history exceeds the context limit, the oldest interactions are truncated, leaving only a sliding window of recent experience. Our setting allows multiple \texttt{play} calls within the same conversation and uses native context compaction when the context limit is reached. This preserves a compressed representation of earlier interactions, allowing information from past turns to remain available for subsequent reasoning.

\textbf{(iii)$\rightarrow$(iv): Add notes.}
We introduce a minimal memory mechanism by adding two note files: \texttt{GUIDE.md} and \texttt{WORKING.md}. \texttt{GUIDE.md} stores reusable knowledge about the game across levels and sessions, while \texttt{WORKING.md} stores temporary state and plans for the current level. When the context approaches its limit, the agent writes a handoff summary in \texttt{WORKING.md} before resuming in a fresh context. These notes remain available across context boundaries, allowing the agent to carry forward its understanding of the game and its current state.

\textbf{(iv)$\rightarrow$(vi): Add VISTA's visual harness.}
Finally, we add VISTA's key visual harness components: lossless visual memory, model-directed inspection (\texttt{inspect}), and exact pixel readout (\texttt{read\_pixels}). The agent receives only the final frame after each action and can decide when to use these tools to revisit its visual memory as it reasons. These additions complete the VISTA harness.

\subsection{Evaluation Metrics}
\label{app:metrics}

\paragraph{ARC-AGI-3.}
\label{app:rhae}
We report Relative Human Action Efficiency (RHAE) using the official ARC-AGI-3 scorer~\citep{arcprize2026arc3,arcprize2026methodology}. For a completed level $\ell$ in game $g$, the level score is
\begin{equation}
S_{\ell,g} = \min\left(1.15,\left(\frac{h_{\ell,g}}{a_{\ell,g}}\right)^2\right),
\end{equation}
where $a_{\ell,g}$ is the agent's action count and $h_{\ell,g}$ is the upper-median action count among first-time human players who completed that level. Unfinished levels receive a score of zero.

For a game with $n$ levels, each level $\ell$ is assigned a weight $w_\ell=\ell$, giving later levels greater weight. If the agent completes the first $k$ levels, the game score is
\begin{equation}
S_g = \min\left(
\frac{\sum_{\ell=1}^{k} w_\ell}{\sum_{\ell=1}^{n} w_\ell},
\frac{\sum_{\ell=1}^{n} w_\ell S_{\ell,g}}{\sum_{\ell=1}^{n} w_\ell}
\right).
\end{equation}
The second term is the weighted average of the level scores, while the first caps the game score by the weighted fraction of levels completed. We report the mean $S_g$ across the 25 public games, multiplied by 100. We additionally report total actions to distinguish the efficiency of runs that receive the same RHAE score.

\paragraph{GameWorld.}
\label{app:gameworld}
\label{app:gameworld-metrics}
GameWorld reports success rate (SR) and progress (PG)~\citep{ouyang2026gameworld}. Success rate is the fraction of tasks completed successfully, while progress averages the benchmark's per-task progress scores, providing credit for partial completion:
\begin{equation}
\mathrm{SR} = \frac{1}{N}\sum_{i=1}^{N}
\mathbb{I}\left[\text{task } i \text{ succeeded}\right],
\end{equation}
\begin{equation}
\mathrm{PG} = \frac{1}{N}\sum_{i=1}^{N} p_i .
\end{equation}
Here, $p_i\in [0,1]$ denotes the highest normalized progress reached during the run for task $i$. We average both metrics over all $N=170$ tasks. Task success and progress are computed by the benchmark's official evaluator from serialized game-state fields.

\paragraph{AI GameStore.}
\label{app:aigs}
Following AI GameStore's evaluation protocol~\citep{ying2026aigamestore}, we normalize each game's raw score $r_g$ by its published human median $m_g$, such that median human performance corresponds to 100. We clip the normalized scores to $[1,10^4]$:
\begin{equation}
  n_g = \operatorname{clip}\!\left(\frac{100 \times r_g}{m_g},\, 1,\, 10^4\right).
\end{equation}
We report the geometric mean of these normalized scores across the ten public games used in our evaluation:
\begin{equation}
\mathrm{Score} = \left(\prod_{g=1}^{10} n_g\right)^{1/10}.
\end{equation}

\paragraph{BabyVision.}
\label{app:babyvision-metrics}
We report accuracy on the evaluated visual tracking questions~\citep{chen2026babyvision}, using exact matching and an LLM judge to assess correctness. Failed runs are counted as incorrect.

\subsection{Agent Tools}

The agent accesses tools only through structured function calls; plain text generated by the model cannot trigger tool execution. In Claude Code, we expose these tools through an MCP server and disable all built-in tools, preventing direct shell or file-system access. In Codex, we register the same tools directly with the app server. In both cases, the agent runs in a sandboxed container without access to environment credentials.

A single host-side controller handles all tool calls. The controller maintains the environment connection, archived frames, \texttt{GUIDE.md}, and \texttt{WORKING.md}, and is the only component allowed to advance the game. It also enforces sequential interaction: a new \texttt{play} call is accepted only after the result of the previous action has been returned to the agent. Each returned image is labeled with its turn and frame index. Claude Code receives these images directly in the MCP result. In our Codex integration, tool outputs are text-only, and the controller provides the corresponding images separately as part of the same turn.

When the context is nearly full, the controller pauses normal tool use until the agent saves the state needed to continue. In Codex, the agent uses \texttt{save\_compact\_checkpoint}; in Claude Code, a pre-compaction hook requests a \texttt{write\_working} call. The agent then resumes in a fresh context containing \texttt{GUIDE.md}, \texttt{WORKING.md}, the action history, and the current frame.

\label{app:tool_prompt}
\definecolor{codeblue}{rgb}{0.25,0.5,0.5}
\definecolor{codefunc}{rgb}{0.85, 0.18, 0.50}
\definecolor{codenetwork}{RGB}{106, 90, 205}
\lstdefinelanguage{PythonFuncColor}{
  language=Python,
  alsoletter={_},
  morekeywords={None,null},
  keywordstyle=\color{blue}\bfseries,
  commentstyle=\color{codeblue},
  stringstyle=\color{orange},
  showstringspaces=false,
  literate=*
    {`play`}{{\color{codefunc}`play`}}{6}
    {`act`}{{\color{codefunc}`act`}}{5}
    {`inspect`}{{\color{codefunc}`inspect`}}{9}
    {`read_pixels`}{{\color{codefunc}`read\_pixels`}}{13}
    {play}{{\color{codefunc}play}}{4}
    {inspect}{{\color{codefunc}inspect}}{7}
    {read_pixels}{{\color{codefunc}read\_pixels}}{11}
    {history}{{\color{codefunc}history}}{7}
    {read_guide}{{\color{codefunc}read\_guide}}{10}
    {write_guide}{{\color{codefunc}write\_guide}}{11}
    {read_working}{{\color{codefunc}read\_working}}{12}
    {write_working}{{\color{codefunc}write\_working}}{13}
    {save_compact_checkpoint}{{\color{codefunc}save\_compact\_checkpoint}}{23}
    {RESET}{{\color{codenetwork}RESET}}{5}
    {ACTION1}{{\color{codenetwork}ACTION1}}{7}
    {ACTION6}{{\color{codenetwork}ACTION6}}{7}
    {ACTION7}{{\color{codenetwork}ACTION7}}{7}
    {GAME_OVER}{{\color{codenetwork}GAME\_OVER}}{9}
    {WIN}{{\color{codenetwork}WIN}}{3}
    {TERMINAL}{{\color{codenetwork}TERMINAL}}{8}
}

\lstset{
  language=PythonFuncColor,
  backgroundcolor=\color{white},
  basicstyle=\fontfamily{cmtt}\fontsize{8pt}{8.9pt}\selectfont,
  columns=fullflexible,
  keepspaces=true,
  breaklines=true,
  captionpos=b,
}

\begin{algorithm}[H]
\caption{\textbf{The VISTA tool interface in pseudocode}. \texttt{play} is the only tool that changes the game environment, note-writing tools update only the agent's stored notes, and all other tools are read-only.}
\begin{minipage}{0.98\linewidth}
\begin{lstlisting}[language=PythonFuncColor]
# ---------------- Action ----------------
play(action, x=None, y=None) -> (frame, n_frames, available_actions)
    # Execute exactly one game action. The action set is defined by each
    # benchmark and game (e.g. ARC-AGI-3: RESET, ACTION1..7, where ACTION6
    # is a click at (x, y); GameWorld and AI GameStore use their own inputs).
    # Returns the final frame. All intermediate frames are archived.

# ------------- Visual memory -------------
inspect(question, views) -> images
    # question: the visual question the views should answer
    # view = {label, turn, frame=final, region=full}, 
    # region = {x, y, w, h}, each view returns one image
    # Retrieve any archived frame, crop the region, and enlarge it

read_pixels(question, views) -> (palette, samples[])
    # question: the visual question the views should answer
    # view = {label, turn, frame=final, region=full}, 
    # region = {x, y, w, h}, each view returns one image
    # Split the region into rows x columns cells and read the pixel value
    # of each cell. Return the values as a list of numbers.

# ---------------- History ----------------
history(view, start_turn=None, end_turn=None, limit=64) -> records[]
    # view = "events":   exact action/result records for turns [start, end)
    # view = "attempts": one summary for each attempt in the current level. An
    # attempt starts when the level is entered or RESET is
    # issued, and ends at GAME_OVER, WIN, or TERMINAL. 

# ----------------- Notes -----------------
# GUIDE.md stores reusable knowledge about the game across levels and sessions.
# WORKING.md stores temporary state and plans for the current level. 
# It is retained across RESETs and context restarts, cleared when the level advances

read_guide() -> text        # read GUIDE.md
write_guide(content)        # replace GUIDE.md
read_working() -> text      # read WORKING.md
write_working(content)      # replace WORKING.md

save_compact_checkpoint(guide, working_memory)
    # Atomically save the state needed to resume after context compaction.
    # Write WORKING.md with working_memory. If guide is not null, also
    # write GUIDE.md; otherwise leave it unchanged.
\end{lstlisting}
\end{minipage}
\end{algorithm}

\subsection{Agent Prompts}
\label{app:prompt}

Each benchmark uses a short prompt following the same template, differing only in benchmark-specific instructions for the action space, task objective, and notes.

\paragraph{ARC-AGI-3.} The same prompt is used for every ARC-AGI-3 game.

\begin{promptbox}[H]
\caption{ARC-AGI-3.}
\begin{minipage}{0.98\linewidth}
\begin{lstlisting}[language=PythonFuncColor, keywordstyle={}]
# Visual game task

Complete the game with as few game actions as possible.

Build and use a compact, revisable model of the game and its current state. Update
it as new evidence changes what is supported.

Before each `play`, briefly state what you expect to see. Afterward, briefly state
all visible changes, expected or not.

Keep concise, durable, revisable game understanding in `GUIDE.md`; use `WORKING.md`
as a scratchpad when useful.
\end{lstlisting}
\end{minipage}
\end{promptbox}

\paragraph{GameWorld and AI GameStore.} GameWorld and AI GameStore use the same environment-neutral prompt. We replace \texttt{play} with \texttt{act} and use the benchmark-provided task objective.
On GameWorld, the system prompt includes the benchmark-provided task prompt verbatim, specifying the objective for each task.
\begin{promptbox}[H]
\caption{GameWorld and AI GameStore.}
\begin{minipage}{0.98\linewidth}
\begin{lstlisting}[language=PythonFuncColor, keywordstyle={}]
# Interactive world task

Pursue the supplied objective.

Build and use a compact, revisable model of the environment and its current state.
Update it as new evidence changes what is supported.

Before each `act`, briefly state what you expect to observe. Afterward, briefly
state all observable changes, expected or not.

Keep concise, durable, revisable understanding in `GUIDE.md`; use `WORKING.md` as a
scratchpad when useful.
\end{lstlisting}
\end{minipage}
\end{promptbox}

\paragraph{BabyVision.}
We use a shorter prompt for BabyVision because each task consists of a single static image. Since tasks are independent, we omit the persistent notes. The agent can gather visual evidence through \texttt{inspect} and \texttt{read\_pixels} before answering.

\begin{promptbox}[H]
\caption{BabyVision.}
\begin{minipage}{0.98\linewidth}
\begin{lstlisting}[language=PythonFuncColor, keywordstyle={}]
# Visual question task

Before answering, gather the evidence the question actually needs from the supplied
image. Use `inspect` to view enlarged regions where detail matters, and
`read_pixels` to measure exact colors where a judgement depends on them. Say what
each view is meant to settle, then check whether it did.

Build a short, revisable account of what the image shows and update it as the
evidence changes what is supported. Answer only when the evidence supports one
answer.
\end{lstlisting}
\end{minipage}
\end{promptbox}

\subsection{Human References}
\label{app:human}

\paragraph{ARC-AGI-3.}
The organizers' study involved 486 participants from the general public with diverse demographics. Participants were not selected for specialized training or skills~\citep{arcprize2026arc3}. Each game was tested by ten participants encountering it for the first time, without task-specific instructions. Testing took place in 90-minute sessions, with a soft time limit of 20 minutes and a hard cutoff of 30 minutes per game. A game was included only if at least two participants independently completed all levels. For each level, the official RHAE human baseline is the upper-median action count among first-time human players who completed that level~\citep{arcprize2026methodology}.

\paragraph{GameWorld.}
GameWorld reports results for two human players, both computer science graduate students, each of whom attempted all 170 tasks under the same budget of 100 actions per task as the agents~\citep{ouyang2026gameworld}. The novice player had no prior exposure to the games or tasks, while the expert player had studied the rules and practiced the controls beforehand. We use the novice player's performance as our reference for first-time human play.

\paragraph{AI GameStore.}
The benchmark's human study recruited 106 participants through Prolific (mean age = 38.81)~\citep{ying2026aigamestore} to play each game over a customized game interface. Each participant was randomly assigned ten games, and was given two minutes per game. We use the median score among participants who played each game as the human reference, normalized to 100 as described above.

\paragraph{BabyVision.}
BabyVision's human evaluation involved 16 adult participants, each holding at least a bachelor's degree, who completed all 388 questions~\citep{chen2026babyvision}. The reported human accuracy is 95.0\% on the Maze subtype and 89.5\% on Connect the Lines. We compute the human reference in Figure~\ref{fig:generalization}, 92.3\%, as the weighted mean of these subtype scores, using the 20 maze and 19 connecting-lines questions in our evaluation.

\section{Additional Results}
\label{app:arc3-details}

\subsection{Run-to-run Variation}
\label{sec:run_variation}

We repeat our main evaluations three times to measure run-to-run variation.
On ARC-AGI-3, VISTA with GPT-5.6 Sol at maximum reasoning effort achieves RHAE scores of 99.00, 98.90, and 99.23 across the three runs, corresponding to $99.04 \pm 0.17$ (mean $\pm$ standard deviation). On GameWorld, the three runs achieve success rates of 64.7\%, 64.7\%, and 60.6\%, and progress scores of 81.0\%, 80.0\%, and 76.9\%, corresponding to $63.33 \pm 2.37$\% success rate and $79.30 \pm 2.14$\% progress. On BabyVision, the three runs achieve 61.5\%, 61.5\%, and 66.7\%, corresponding to $63.23 \pm 3.00$\%.

For AI Game Store, the three runs obtain aggregate scores of 119.99, 148.58,
and 140.06. We follow the benchmark's official multi-run evaluation protocol
rather than averaging these three final aggregate scores directly: for each
game, we first average its raw score across the three runs, and then apply the
benchmark's normalization and aggregation procedure to these per-game means.

\subsection{Per-game Results on ARC-AGI-3}
\label{app:pergame}

Table~\ref{tab:pergame} reports per-game results for VISTA with Claude Opus 5.0 and GPT-5.6 Sol, alongside the first-time human action reference.

\begin{table}[t]
\caption{\textbf{Per-game results on the 25 public ARC-AGI-3 games.} We evaluate VISTA with Claude Opus 5.0 and GPT-5.6 Sol. For each game, human actions are the sum of the per-level human baselines, each defined as the upper-median action count among first-time human players who completed that level. The action ratio is the agent's total actions divided by this reference. Claude Opus 5.0 achieves an RHAE score of 100.00 on every game. GPT-5.6 Sol completes every game and achieves a mean RHAE score of 99.00.}
\label{tab:pergame}
\begin{center}
\footnotesize
\setlength{\tabcolsep}{4pt}
\begin{tabular}{lrrrrrrr}
\toprule
& & \multicolumn{3}{c}{\textbf{Claude Opus 5.0}} & \multicolumn{3}{c}{\textbf{GPT-5.6 Sol}} \\
\cmidrule(lr){3-5} \cmidrule(lr){6-8}
\textbf{Game} & \textbf{Human actions} & \textbf{Score} & \textbf{Actions} & \textbf{Ratio} & \textbf{Score} & \textbf{Actions} & \textbf{Ratio} \\
\midrule
ar25 & 748 & 100.00 & 256 & 0.34$\times$ & 100.00 & 260 & 0.35$\times$ \\
bp35 & 651 & 100.00 & 521 & 0.80$\times$ & 100.00 & 534 & 0.82$\times$ \\
cd82 & 171 & 100.00 & 112 & 0.65$\times$ & 100.00 & 100 & 0.58$\times$ \\
cn04 & 789 & 100.00 & 225 & 0.29$\times$ & 100.00 & 200 & 0.25$\times$ \\
dc22 & 1{,}228 & 100.00 & 547 & 0.45$\times$ & 100.00 & 560 & 0.46$\times$ \\
ft09 & 208 & 100.00 & 75 & 0.36$\times$ & 100.00 & 77 & 0.37$\times$ \\
g50t & 879 & 100.00 & 338 & 0.38$\times$ & 100.00 & 486 & 0.55$\times$ \\
ka59 & 730 & 100.00 & 303 & 0.42$\times$ & 100.00 & 396 & 0.54$\times$ \\
lf52 & 1{,}339 & 100.00 & 1{,}008 & 0.75$\times$ & 100.00 & 936 & 0.70$\times$ \\
lp85 & 388 & 100.00 & 104 & 0.27$\times$ & 100.00 & 97 & 0.25$\times$ \\
ls20 & 776 & 100.00 & 458 & 0.59$\times$ & 75.03 & 1{,}135 & 1.46$\times$ \\
m0r0 & 1{,}107 & 100.00 & 219 & 0.20$\times$ & 100.00 & 224 & 0.20$\times$ \\
r11l & 233 & 100.00 & 75 & 0.32$\times$ & 100.00 & 93 & 0.40$\times$ \\
re86 & 1{,}255 & 100.00 & 628 & 0.50$\times$ & 100.00 & 582 & 0.46$\times$ \\
s5i5 & 638 & 100.00 & 254 & 0.40$\times$ & 100.00 & 293 & 0.46$\times$ \\
sb26 & 213 & 100.00 & 125 & 0.59$\times$ & 100.00 & 124 & 0.58$\times$ \\
sc25 & 350 & 100.00 & 153 & 0.44$\times$ & 100.00 & 228 & 0.65$\times$ \\
sk48 & 1{,}070 & 100.00 & 413 & 0.39$\times$ & 100.00 & 605 & 0.57$\times$ \\
sp80 & 518 & 100.00 & 119 & 0.23$\times$ & 100.00 & 184 & 0.36$\times$ \\
su15 & 361 & 100.00 & 99 & 0.27$\times$ & 100.00 & 123 & 0.34$\times$ \\
tn36 & 317 & 100.00 & 143 & 0.45$\times$ & 100.00 & 225 & 0.71$\times$ \\
tr87 & 414 & 100.00 & 140 & 0.34$\times$ & 100.00 & 209 & 0.50$\times$ \\
tu93 & 462 & 100.00 & 193 & 0.42$\times$ & 100.00 & 197 & 0.43$\times$ \\
vc33 & 447 & 100.00 & 182 & 0.41$\times$ & 100.00 & 272 & 0.61$\times$ \\
wa30 & 1{,}843 & 100.00 & 612 & 0.33$\times$ & 100.00 & 986 & 0.53$\times$ \\
\midrule
\textbf{Total (25 games)} & \textbf{17{,}135} & \textbf{100.00} & \textbf{7{,}302} & \textbf{0.43$\times$} & \textbf{99.00} & \textbf{9{,}126} & \textbf{0.53$\times$} \\
\bottomrule
\end{tabular}
\end{center}
\end{table}

\subsection{Interaction trajectories}
\label{app:trajectories}

\begin{figure}[t]
\centering
\includegraphics{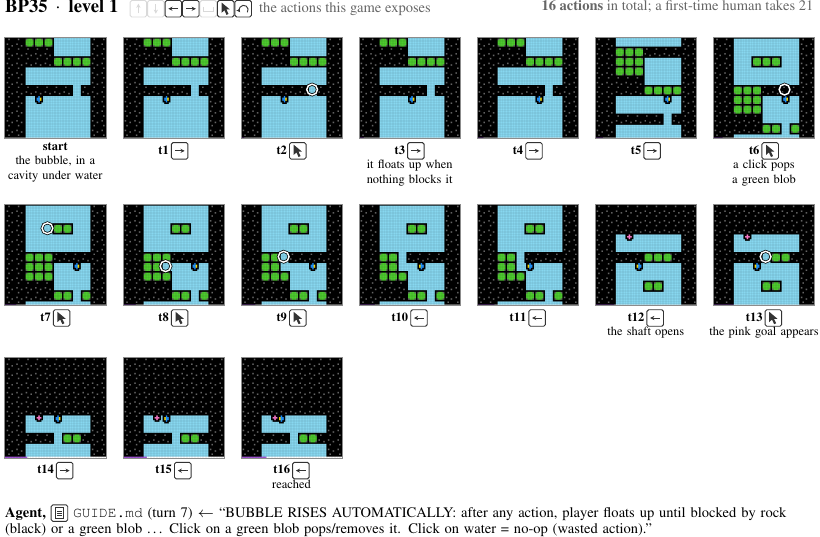}
\caption{\textbf{A full VISTA trajectory on an ARC-AGI-3 task} (ID BP35), one frame per action (level 1). The player begins under upward gravity and can move left or right. Clicking a green block removes it, allowing the player to clear obstacles and navigate through otherwise blocked passages. The goal is to reach the pink object. Each frame shows the state after the labeled action; rings indicate clicks. Read left to right, top to bottom.}
\label{fig:arc3-traj-bp35-l1}
\end{figure}

\begin{figure}[ht]
\centering
\includegraphics{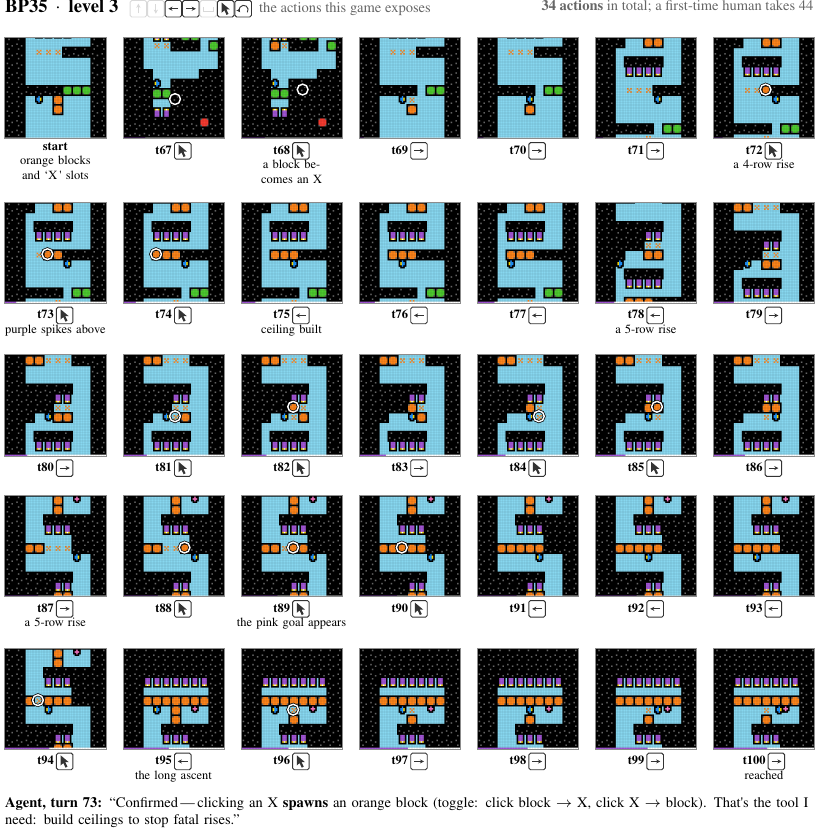}
\caption{\textbf{A full VISTA trajectory on an ARC-AGI-3 task} (ID BP35), one frame per action (level 3). Level 3 introduces an orange block that disappears when clicked and reappears when clicked again. The agent can remove it to pass through a blocked path, or restore it to prevent the player from rising into the deadly purple spikes.}
\label{fig:arc3-traj-bp35-l3}
\end{figure}

\begin{figure}[t]
\centering
\includegraphics{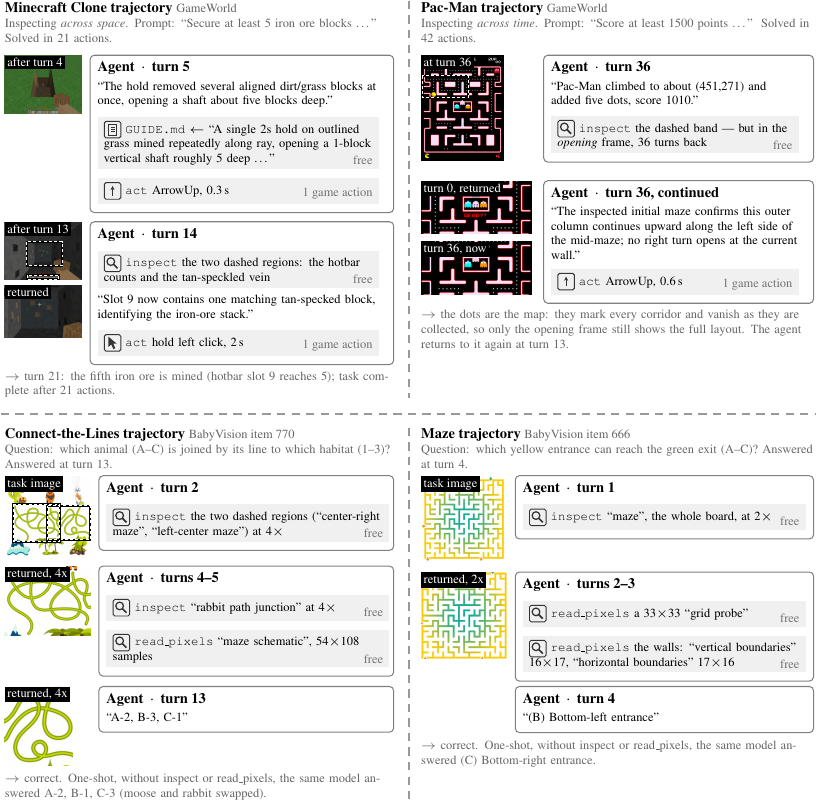}
\caption{\textbf{Four VISTA trajectories beyond ARC-AGI-3} with GPT-5.6 Sol. The top row shows GameWorld games, and the bottom row shows BabyVision tasks. \textit{Top left}: in Minecraft Clone, the agent inspects across space, magnifying the hotbar and ore vein to distinguish iron from stone. \textit{Top right}: in Pac-Man, it inspects across time, revisiting the current and initial frames to view the full maze. \textit{Bottom left}: in Connect the Lines, the agent magnifies two regions to trace a path that is difficult to follow at the supplied image resolution, then reads exact pixel values before answering. \textit{Bottom right}: in Maze, it renders the board at $2\times$ scale and then reads the wall grid as pixel values.}
\label{fig:beyond}
\end{figure}

We show additional VISTA trajectories on BP35 from ARC-AGI-3 in Figures~\ref{fig:arc3-traj-bp35-l1} and~\ref{fig:arc3-traj-bp35-l3}. We also show trajectories from visual environments beyond ARC-AGI-3 in Figure~\ref{fig:beyond}, including two interactive games from GameWorld and two static visual reasoning tasks from BabyVision.

